\documentclass[11pt,a4paper]{article}
\usepackage[hyperref]{acl2020}
\usepackage{times}
\usepackage{url}
\usepackage{latexsym}
\usepackage{xcolor}
\usepackage{hyperref}
\usepackage{subcaption}
\usepackage{graphicx}
\usepackage{amsmath}
\usepackage{array}
\usepackage{tipa}
\usepackage{booktabs}
\usepackage[none]{hyphenat} % remove hyphens
\usepackage{enumitem}
\usepackage{array} % table cell alignment
\usepackage{makecell}
\usepackage{microtype}

\newcommand{\rowgroup}[1]{\hspace{-1em}#1}

\aclfinalcopy % Uncomment this line for the final submission
\def\aclpaperid{69} %  Enter the acl Paper ID here

\title{Tuiteamos o pongamos un tuit? Investigating the Social Constraints of Loanword Integration in Spanish Social Media}

\author{Ian Stewart\thanks{Work completed at Georgia Institute of Technology.} \\
  University of Michigan \\
  \texttt{ianbstew@umich.edu} \\ \And
  Diyi Yang \\
  Georgia Institute of Technology \\
  \texttt{diyi.yang@cc.gatech.edu} \\ \AND 
  Jacob Eisenstein \\
  Google Research \\
  \texttt{jeisenstein@google.com}}

\date{}

\begin{document}
\maketitle

\begin{abstract}
Speakers of non-English languages often adopt loanwords from English to express new or unusual concepts.
While these loanwords may be borrowed unchanged, speakers may also integrate the words to fit the constraints of their native language, e.g. creating Spanish \emph{tuitear} from English \emph{tweet}.
Linguists have often considered the process of loanword integration to be more dependent on language-internal constraints, but sociolinguistic constraints such as speaker background remain only qualitatively understood.
We investigate the role of social context and speaker background in Spanish speakers' use of integrated loanwords on social media.
We find first that newspaper authors use the integrated forms of loanwords and native words more often than social media authors, showing that integration is associated with formal domains.
In social media, we find that speaker background and expectations of formality explain loanword and native word integration, such that authors who use more Spanish and who write to a wider audience tend to use integrated verb forms more often.
This study shows that loanword integration reflects not only language-internal constraints but also social expectations that vary by conversation and speaker.
\end{abstract}

\section{Introduction}
Languages exchange loanwords constantly as multilingual people adopt words from other languages to express themselves in their native language~\citep{haspelmath2009}.
The English word \emph{tweet} has been adopted into many other languages following the success of Twitter, e.g. producing the Spanish verb \emph{tuitear}.
One form of adoption is known as \emph{integration} by which a speaker adapts the loanword to the underlying grammar of the language, e.g. adding the Spanish verb ending \emph{-ear} to the loanword \emph{tweet} to help the word adhere to Spanish grammar~\citep{poplack2012}.
Speakers may choose to use loanwords with the prescriptively correct form, in this case adding verbal morphology, or with less standard forms, in this case using a paraphrase such as \emph{send a tweet}.
We show several examples of this alternation in \autoref{tab:loanword_summary_stats}.
To further the theoretical understanding of the process of loanword integration, this work assesses this process from a speaker's perspective.

\begin{table}[t!]
    \centering
    \begin{tabular}{b{1.25cm} b{4.5cm} b{1cm}}
        Loanword & Verbs & Count \\ \toprule
        Connect & \emph{conectear}, \emph{hacer un conexión} & 7785 \\
        Like & \emph{likear}, \emph{dar un like} & 5666 \\
        Stalk & \emph{stalkear}, \emph{ser un stalker} & 5455 \\
        Flash & \emph{flashear}, \emph{hacer flash} & 4521 \\
        Ship & \emph{shippear}, \emph{hacer ship} & 4079 \\
    \end{tabular}
    \caption{Top 5 most frequent loanwords on social media and corresponding verb forms.}
    \label{tab:loanword_summary_stats}
\end{table}

Researchers have often studied the process of loanword adoption and integration from a language-internal perspective, such as phonological constraints on loanword use~\citep{kang2011}.
However, loanwords also carry \emph{social meaning}~\citep{levendis2019} that relates to formality and standard language norms, and speakers may have their own intuitions about the ``correct'' way to use a loanword.
Therefore, a speaker's background, such as their multilingual knowledge~\citep{poplack1988}, and the social context of a conversation~\citep{levari2014experiment} may also play a role in the integration of loanwords.
Such social and behavioral factors may also help explain the long-term \emph{acceptance} of loanwords into a language~\citep{chesley2010,zenner2012}.
To that end, we leverage multilingual data from social media to assess the speaker-level factors that underlie loanword integration.

Our study provides the following contributions:
\begin{itemize}
\setlength\itemsep{0pt}
\setlength\parskip{0pt}
    \item We first collect verb forms for a variety of English loanwords related to technology and social life online, as well as similar \emph{control} pairs for native Spanish verbs (\autoref{subsec:identifying_loanwords},~\autoref{subsec:native_verbs}).
    \item To test for the effect of formality, we compare the rate of integrated verb use for loanwords and native verbs between social media posts and newspaper articles (\autoref{sec:results_domain_comparison}).
    We find that loanwords and native verbs are integrated at a higher rate in newspaper articles, suggesting that integration is associated with more formal language registers.
    \item Drawing on this finding, we test the role of different contextual and speaker-background factors as they explain the choice to use integrated verbs for loanwords (\autoref{sec:variable_summary}, \autoref{sec:results_speaker_comparison}).
    With regression analysis on social media data, we show that speaker background plays a large role: Latin American speakers and high-Spanish speakers tend to choose integrated verbs for loanwords and native words.
    We also find that the context of a post explains integration, because posts with a larger presumed audience have higher rates of integration.
    Lastly, we find several points of divergence between loanwords and native verbs, suggesting some differences in social perception of the word groups.
\end{itemize}
\section{Related work}

Loanword integration has mainly been studied from the perspective of \emph{pronunciation}, i.e. whether a loanword adheres to the phonology of the source or target language~\citep{kang2011}.
Speakers may have to choose between different valid pronunciations, e.g. pronouncing the word \emph{Iraq} with an American English ``short-A'' (/\textipa{I\*r\ae{}k}/) or an Arabic ``long-A'' (/\textipa{I\*rAk}/)~\citep{hall2010}.
Traditional studies of loanword integration relied on sociolinguistic interviews and elicitation, which often lack spontaneous loanword use~\citep{poplack1988}.
With the growing availability of large-scale written corpora, researchers have tracked the adoption of loanwords over time, particularly English loanwords into other languages~\citep{chesley2010,garley2012,zenner2012}.
Such large-scale corpora also allow researchers to track \emph{morphological} integration~\citep{coats2018,kilgarriff2010}, which is a word's ability to combine with bound morphemes from the target language (e.g. \emph{tuitear} [``to tweet''] = \emph{tuit} [``tweet''] + \emph{-ear} [VERB.INF]).
We continue this line of work and study the role of contextual and speaker-background factors in loanword integration.
This helps test theories related to multilingual decisions~\cite{poplack2020} and how loanwords are collectively adopted into a language~\citep{levendis2019}.

The loanword integration process relates partly to structure: if the source and target language are similar (e.g. Italian and Spanish) then a speaker may have little difficulty in integrating the loanword~\citep{boersma2009,peperkamp2004}.
However, a speaker's decision to integrate a loanword also depends on the speaker's prior experiences and the social context of the conversation~\citep{wohlgemuth2009}.
For one, the choice of using an integrated loanword depends on the speaker's own background with the source language~\citep{poplack1988} and their willingness to uphold linguistic standards for the loanword.
In addition, the process of loanword integration may be related to the \emph{domain} of speech, as some writing domains such as newspapers have strong norms~\citep{biber2019} and therefore may prefer the formal version of the loanword.
Lastly, the social expectations of a given \emph{conversation} may convince a speaker to use the integrated form~\citep{levari2014experiment}, e.g. if their listeners are expecting a less formal response and therefore a non-integrated loanword.
While some work has tested both linguistic and social constraints on the integration of loanwords~\citep{garley2014,sanchez2005}, linguists generally lack access to speech across a variety of speakers and social contexts.
This work addresses the social meaning of loanwords by drawing on the rich speaker-level data available from social media.
\section{Data}

\subsection{Identifying Loanwords}
\label{subsec:identifying_loanwords}

The use of a loanword is considered distinct from code-switching (switching between languages), because a loanword is produced in isolation within the ``matrix'' language~\citep{poplack1988,cacoullos2003}.
This study concerns the alternation between integrated verbs, i.e. those in which the loanword has been morphologically integrated into the language (\emph{tuitear} ``to tweet'') and light verbs, i.e. phrases in which the loanword is used as a noun (\emph{poner un tweet} ``to send a tweet'').
We seek light verb phrases that are semantically similar to the integrated verbs, to avoid possible confounds on the choice between forms.

The list of loanword integrated verbs was identified from two resources: Wiktionary and social media.
We first collected all verbs on Spanish-language Wiktionary that are English-origin loanwords and end in one of the standard verb suffixes (\emph{-(e)ar}).\footnote{Accessed 1 Jan 2020: \url{https://es.wiktionary.org/wiki/Categoria:ES:Palabras_de_origen_ingles}.}
Using a sample of Reddit and Twitter data,\footnote{Data sample of Spanish-language posts ranges from 1 July 2017 to 30 June 2019. For Reddit this includes all comments ($\sim$560,000), for Twitter this includes a 1\% sample from the Twitter stream ($\sim$110,000,000).} we collected all words in Spanish-language posts tagged using \texttt{langid}~\cite{lui2012} that match the structure \texttt{ENGLISH\_WORD} + \emph{-(e)ar},\footnote{English words collected from a standard spellcheck dictionary and filtered to exclude words shorter than $n=4$ characters. Accessed 1 Nov 2019: \url{http://wordlist.aspell.net/dicts/}.} under the assumption that most loanword verbs use the \emph{-(e)ar} conjugation~\citep{rodney2012}.
From the combined set of verbs, we removed all cases of ambiguity, e.g. \emph{plantar}, which can be formed by English \emph{plant} + \emph{-ar}, is also a native Spanish word.

For each loanword, we identified a corresponding light verb phrase with a meaning similar to the integrated form.
Spanish has a closed class of light verbs used to form phrases with nouns~\citep{buckingham2013}, such as \emph{tomar} (``take'') in \emph{tomar un viaje} (``take a vacation'').
We used dictionary definitions from Wiktionary and WordReference to identify valid light verb forms, and we queried the internet for the remaining loanwords to determine their validity (e.g. comparing search results for \emph{hacer un tweet} versus \emph{poner un tweet}).
We validated the loanword pairs with Spanish linguistics experts familiar with the process of loanword integration.
The experts removed several loanwords that may have been considered native words by Spanish speakers.\footnote{E.g., Spanish speakers may not consider \emph{flipar} (``to flip'') to be a loanword due to its older status.}

This process yielded 120 integrated and light verb pairs that we used to define the dependent variable of the study, i.e. integrated verb use vs. light verb use.
We show examples of the most frequent loanword and light verb pairs in \autoref{tab:loanword_summary_stats}.
Many of the words identified relate to technology and online behavior (e.g. \emph{likear} ``to like (on social media)''), which represents a sample bias.
Because we study loanword use specifically on Twitter, it is likely that the loanwords here relate more to the interests of the platform community rather than the general population.

\subsection{Identifying Native Verbs}
\label{subsec:native_verbs}

Studying loanwords in isolation can yield interesting results, but we must also determine whether the patterns of usage reflect constraints on Spanish verbs in general~\citep{wichmann2008}.
To address this concern, we collect an additional set of verbs that are native to Spanish.

We first identified light verb constructions from several grammar blogs and dictionaries,\footnote{E.g. ``support verbs'' mentioned here, accessed 1 Jan 2020: \url{https://comunicarbien.wordpress.com/2011/08/06/verbos-de-apoyo/}.} and generated the corresponding integrated verb by adding a standard verb suffix to the noun phrase and verifying with a dictionary.\footnote{E.g. for the light verb construction \emph{tomar un viaje} (``to take a trip'') with the noun \emph{viaje}, we generated the integrated verb \emph{viajar} (``to travel'').}
This process yielded 49 pairs of native integrated and light verbs that serve as a baseline to compare with loanword use.
We extracted all uses of these native verbs from the set of loanword-using authors mentioned above.
As shown in \autoref{tab:native_verb_summary_stats}, the native verbs occur more frequently than the loanword verbs, which compensates for the fact that we have fewer word types for native verbs.

\begin{table}[t!]
    \centering
    \begin{tabular}{b{1.5cm} b{4cm} b{1cm}}
        Native word & Verbs & Count \\ \toprule
        Dream & \emph{soñar}, \emph{tener un sueño} & 39,392 \\
        Buy & \emph{comprar}, \emph{hacer la compra} & 36,337 \\
        End & \emph{terminar}, \emph{poner término} & 34,234 \\
        Use & \emph{usar}, \emph{hacer uso} & 30,834 \\
        Test & \emph{probar}, \emph{poner a prueba} & 29,930 \\
    \end{tabular}
    \caption{Top 5 most frequent native word pairs and corresponding verb forms on social media.}
    \label{tab:native_verb_summary_stats}
\end{table}

The complete list of loanwords and native verbs is provided in \autoref{sec:appendix} for replicability and for linguists to build upon in future work.

\subsection{Collecting Loanword Author Data}
\label{subsec:loanword_data}

For our social media data, we collect posts from a 1\% Twitter archive sample of Spanish-language posts, ranging from 1 July 2017 to 30 June 2019.
We match all original (non-RT) posts that contain at least one loanword verb form, either in the integrated form or light verb form.\footnote{We searched for the most frequently inflected forms of each verb, which include all forms of indicative present, simple past and imperfect. We also remove all verb forms that are ambiguous: e.g. the verb \emph{acceso} (``I access'') has the same spelling as the noun \emph{acceso} (``access'').}
This yields roughly 87,000 posts from 80,000 unique authors over  the period of study, from which roughly 23,000 posts from 20,000 authors were used in the regression, after filtering for available variables described in \autoref{sec:variable_summary}.

Next, we collect all available prior posts from these loanword authors using both the original archive sample (2017-2019) and from the authors' full timelines (2014-2019).\footnote{Collected in Mar 2020.}
We recovered roughly 10 million posts from the authors (about 100 extra posts per author) from which we extracted native verb use and speaker background variables for analysis (see \autoref{tab:all_regression_variables}).

\subsection{Extracting Speaker-Level Variables}

\label{sec:variable_summary}

\begin{table*}[]
\centering
\small
\begin{tabular}{>{\quad}p{2.5cm} p{1.5cm} p{3cm} >{\raggedright}p{3.25cm} p{3.25cm}}
\toprule
\rowgroup{\textbf{Variable type}} & \textbf{Name} & \textbf{Description} & \textbf{Mean / distribution} & \\
\cmidrule{4-5}
\rowgroup{\textbf{Formality}} & & & \textbf{Loanword posts} & \textbf{Native word posts}\\

Post content & Hashtag & Whether post contains a hashtag. & 8.1\% & 6.6\% \\
~ & Mention & Whether post contains an @-mention. & 35.2\% & 7.4\% \\
~ & Post length & Length of post in characters, excluding the verb phrase. & 88 & 131 \\ \hline
\rowgroup{\textbf{Background}} & ~ & ~ & \multicolumn{2}{l}{\textbf{All authors}} \\
Posting behavior & Activity & Mean posts per day. & \multicolumn{2}{l}{8.5} \\
~ & Content re-sharing & Percent of prior posts that are retweets. & \multicolumn{2}{l}{35.2\%} \\
~ & Link sharing & Percent of prior posts that contain a URL. & \multicolumn{2}{l}{0.5\%} \\
Location & Location & Author's geographic region based on self-reported location. & \multicolumn{2}{p{6.5cm}}{54.6\% UNK, 34.7\% Latin America, 7.0\% Europe, 2.7\% US, 0.9\% Other} \\
Language & Language type & Percent of prior posts written in Spanish. & \multicolumn{2}{p{6cm}}{83.8\% high Spanish, 15.5\% medium Spanish, 0.7\% low Spanish} \\
~ & Verb use & Percent of prior native verb posts that contain an integrated verb. & \multicolumn{2}{l}{95.4\%} \\ \bottomrule
\end{tabular}
\caption{Summary of all social media variables used in study.}
\label{tab:all_regression_variables}

\end{table*}

For the speaker-level analysis, we seek to assess the relative importance of several author-level and post-level factors in explaining loanword integration.
Following prior work in loanword use, we investigate factors related to \textbf{formality}~\cite{biber2019} and aspects of \textbf{speaker background}~\cite{poplack2012} that reflect support for language standards.
We therefore use the following metrics to predict verb integration.

\begin{itemize}[noitemsep]
    \setlength\parskip{0pt}
    \item \textbf{Formality}:
    \begin{itemize}[leftmargin=1.1em]
        \item \textbf{Post features}: 
        First, we approximate a post's intended \emph{audience} by marking the presence of a hashtag (larger audience) and the presence of an @-mention (smaller audience).
        We also use the length of a post --- excluding the verb phrase --- to identify posts that are longer and therefore potentially more formal, following prior work in perceptions of formality in online communication~\cite{chhaya2018,pavlick2016}.
    \end{itemize}
    \item \textbf{Speaker background}:
    \begin{itemize}[leftmargin=1.1em]
        \item \textbf{Posting behavior}: 
        Authors who post frequently may have more extensive knowledge of linguistic norms online and therefore adhere to the standard integrated verb form.
        For this metric, we extract the author's mean number of prior posts per day.
        In addition, authors who share more content online may also be more connected to online norms and may therefore adopt the more standard verb form.
        We compute an author's rate of sharing as (1) the percentage of prior posts that contain a URL and (2) the percentage of prior posts that are retweets.
        \item \textbf{Location}: The Spanish dialects spoken in Latin America have diverged significantly from Castilian Spanish~\citep{lipski1994}, which may result in different patterns of loanword adoption. 
        We identify authors' location\footnote{Following prior work~\citep{kariryaa2018}, we use an author's self-reported location in their profile as a location marker. We define an author as a resident of a particular country based on the presence of unambiguous country, state or city keywords in their profile location.} at the region level: Latin America, US, Europe, or other.\footnote{We acknowledge the considerable diversity of Spanish dialects spoken in Latin America~\cite{buckingham2013}, but we use the level of region in our analysis to avoid data sparsity.}
        \item \textbf{Language use}: Bilingual speakers may be more likely to use the light verb forms of the loanwords~, because bilingual speakers often use paraphrases to address unfamiliar concepts~\citep{jenkins2003} and may perceive light verb constructions differently~\cite{dougruoz2014}.
        We tag the authors' prior posts using \texttt{langid},\footnote{We filter to posts with a confidence score above 90\% to reduce likelihood of code-switching.} and compute the rate of Spanish use for all authors who have written at least 5 posts.
        We then bin language use under the assumption that language use may not be linear.
        Authors who use exclusively Spanish (100\%) are assumed to be ``strict'' monolingual speakers as compared to more ``relaxed'' bilingual (0-50\%) or mid-range bilingual (50-100\%) speakers.
        
        In addition to language choice, speakers who use more integrated native verbs may also use more integrated forms for loanwords.
        We compute the authors' rate of prior integrated verb use as the number of integrated native verb tokens (\autoref{subsec:native_verbs}) normalized by the total number of native verb tokens.
    \end{itemize}
\end{itemize}

All variables in the social media data are summarized in \autoref{tab:all_regression_variables}.
Note that we choose not to analyze individuals' gender and age due to the relative difficulty of extracting such information from social media data, particularly in non-English contexts~\citep{wang2019}.
\section{Results}

\subsection{Domain Differences in Loanword Integration}
\label{sec:results_domain_comparison}

\begin{figure*}
    \centering
    \begin{subfigure}{0.48\textwidth}
    \includegraphics[width=\textwidth]{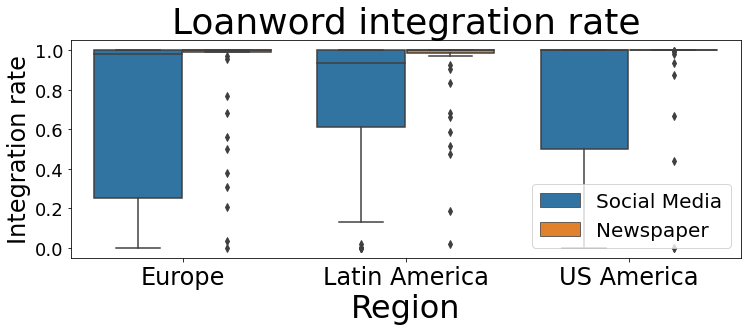}
    \label{fig:integrated_verb_rate_loanwords_newspapers_social_media}
    \end{subfigure}
    \begin{subfigure}{0.48\textwidth}
    \includegraphics[width=\textwidth]{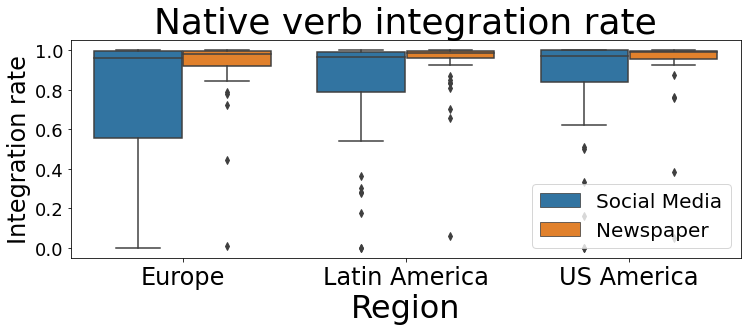}
    \label{fig:integrated_verb_rate_native_words_newspapers_social_media}
    \end{subfigure}
    \caption{Integrated verb use across social media text (blue/left) and newspaper text (orange/right). Each unit is the ratio of integrated verb use for a single word type.}
    \label{fig:integrated_verb_use_newspaper_social_media}
\end{figure*}
The first hypothesis to test concerns the role of domain.
As newspapers are generally considered more formal than social media~\citep{biber2019,pavlick2016}, we expect that loanwords and native verbs to be produced with the presumably more formal integrated forms.
\\

\textbf{H1}: Writers in a more formal domain will tend to use the integrated form of loanwords at a higher rate than writers in a less formal domain.
\\

To test this hypothesis, we collect data from a corpus of Spanish language newspapers from 21 different Spanish-speaking countries and regions.\footnote{News On the Web Spanish, approximately 7 billion tokens over 25 million documents, accessed May 2020: \url{https://www.corpusdelespanol.org/now/}.}
We collect the 50 most frequent loanword pairs and native verb pairs from the social media data and compute their raw frequencies in the newspaper data.
For each pair of integrated verb and light verb, we compute the rate of integrated verb use as the normalized frequency of the integrated verb.
Formally, for a word base $w$, the set of all integrated verb forms $\mathcal{W}_{i,w}$, and the set of all light verb forms for the word $\mathcal{W}_{l,w}$, the rate of integrated verb use $I_w$ is defined as:

$$I_w = \frac{\sum_{w_i \in \mathcal{W}_{i,w}} \text{count}(w_i)}{\sum_{w' \in W_{i,w} \cup W_{l,w}} \text{count}(w')}$$

We show the rates of integration across domains and locations in \autoref{fig:integrated_verb_use_newspaper_social_media}.
The first key finding is that the rate of integration is not significantly different for newspapers across locations, despite known dialect differences across regions.
In addition, we see that for loanwords both social media and newspapers favor the integrated form over the light verb form, in correspondence with the expected ``hierarchy'' of loanword adaptation that places light verbs below integration~\citep{wohlgemuth2009}.
With respect to \textbf{H1}, we see that newspaper writers consistently use the integrated form of loanwords and native verbs more frequently than the social media authors.
Loanwords are integrated at a mean per-word rate of $91\%$ in the newspapers as compared to $82\%$ in social media, while native verbs have a rate of $93\%$ in the newspapers and $82\%$ in social media.\footnote{Both cases had a significant difference with $p<0.01$ by Wilcoxon's signed-rank test.}
We show in \autoref{fig:integrated_verb_use_newspaper_social_media} that this difference holds across all regions.\footnote{We find $p < 0.05$ across all location pairs except loanwords in US America and native verbs in Latin America, by Wilcoxon's test with Bonferroni correction.}

The consistent difference between social media and newspaper writing suggests that the domain of newspaper writing has more formal standards with respect to the use of both loanwords and native words~\citep{geeraerts2003}.
Such consistency may reflect differences in how newspaper writers are expected to cover emerging phenomena such as new loanwords.
A newspaper writer might be encouraged to use the formal version of a newer loanword to maximize the likelihood of their readers' understanding the word~\cite{iwasaki1994,llopis2009}.
To investigate this in more detail, we show the loanwords with the highest absolute difference in integration rate across social media and newspapers in \autoref{tab:integrated_rate_differences_social_media_newspapers}.
The loanwords that are integrated more often in newspapers seem to be relatively newer and possibly related more to online social media activity (e.g. \emph{block}, \emph{hype}), while the loanwords that are integrated more often on social media seem to be somewhat older and relevant to a wider range of activities (e.g. \emph{host}, \emph{rock}).
This finding about domain reinforces the \emph{social} meaning of loanword use, which informs the following speaker-level analysis.

\begin{table}[t!]
    \centering
    \begin{tabular}{l r r r} 
        Word & $I_{w, \text{social media}}$ & $I_{w, \text{newspaper}}$ & $\Delta \: I_{w}$ \\ \toprule
        zap   & 0.179 & 1.000 & -0.821 \\
        block    & 0.153 & 0.857 & -0.704 \\
        hype    & 0.393 & 0.995 & -0.602 \\
        link    & 0.335 & 0.872 & -0.536 \\
        like & 0.115 & 0.649 & -0.534 \\
        ... & ... & ... & ... \\
        pitch & 0.998 & 0.988 & 0.011 \\
        host & 0.990 & 0.972 & 0.018 \\
        google & 0.561 & 0.531 & 0.030 \\
        rock & 0.787 & 0.648 & 0.139 \\
        DM & 1.000 & 0.120 & 0.880 \\ \bottomrule
    \end{tabular}
    \caption{Loanwords with biggest differences in integration between newspaper and social media.}
    \label{tab:integrated_rate_differences_social_media_newspapers}
\end{table}

\subsection{Speaker-level factors in loanword integration}
\label{sec:results_speaker_comparison}

We now turn to speaker-level data to assess the relative impact of different social factors in the use of integrated loanwords. 
If integrated verbs are considered more formal than light verbs (\autoref{sec:results_domain_comparison}), then we expect factors relevant to formality and speech standards to predict integrated verb use for both loanwords and native verbs:
\\

\textbf{H2}: Speakers in social contexts that prefer formal language standards, and with backgrounds that support more standard language use, will tend to use integrated loanwords.
\\

We use logistic regression to predict the use of an integrated verb (1/0) for a given loanword or native word, using different subsets of post-level and speaker-level features specified in \autoref{sec:variable_summary}.
We add fixed effects for all sufficiently frequent authors and word types.\footnote{All authors and words with a count less than N=5 were assigned to a \texttt{RARE} category to avoid sparsity.}
To avoid overfitting the fixed effect variables, we choose an L2 weight for ridge regression, in order to maximize likelihood on held-out data.\footnote{Weight selected from grid search to maximize held-out likelihood on a 10\% test split of the data, for each separate regression.}
For the default values of categorical variables in the regression, we specify ``Unknown'' for author location and ``low Spanish'' for prior language use.
All scalar variables (post length, post activity, content sharing, link sharing, native integrated verb use) were log-transformed and Z-normalized before regression.

We show the social media regression results in \autoref{tab:integrated_verb_regression_combined}.
The following significant results emerge from the analysis.

\begin{table*}
\centering
\renewcommand{\arraystretch}{1.05} 
\begin{tabular}{>{\quad}p{4cm} p{3cm}  >{\raggedright}p{1.25cm} l >{\raggedright}p{1.25cm} l}
\toprule
 &  & \multicolumn{2}{c}{Native words} & \multicolumn{2}{c}{Loanwords} \\ 
\rowgroup{Variable type} & Variable & $\beta$ & S.E. & $\beta$ & S.E. \\ \midrule
 & Intercept & 2.572* & 0.030 & 1.376* & 0.234 \\ 
\rowgroup{\textbf{Formality}} & ~ & ~ & ~ & ~ & ~ \\
Post features & Has hashtag & 0.099* & 0.010 & 0.079 & 0.026 \\
 & Has mention & -0.050* & 0.009 & -0.087* & 0.015 \\
 & Post length & \textbf{-0.046}* & 0.002 & \textbf{0.051}* & 0.008 \\ 
\rowgroup{\textbf{Background}} & ~ & ~ & ~ & ~ & ~ \\
Author behavior & Post activity & 0.006 & 0.003 & -0.034 & 0.011 \\
 & URL sharing & \textbf{-0.015}* & 0.003 & \textbf{0.024}* & 0.010 \\
 & RT sharing & 0.025* & 0.003 & -0.010 & 0.009 \\ 
 Location & Latin America & 0.133* & 0.005 & 0.228* & 0.016 \\
& Europe & -0.223* & 0.010 & -0.367* & 0.033 \\
 & US & 0.008 & 0.015 & -0.143 & 0.048 \\
 & Other & 0.171* & 0.025 & -0.193 & 0.082 \\
Language & High Spanish & 0.606* & 0.031 & 0.589* & 0.110 \\
 & Medium Spanish & 0.687* & 0.030 & 0.424* & 0.107 \\
 & Integrated verb use & ~ & ~ & -0.006 & 0.007 \\ 
 \midrule
\rowgroup{Sample size} &  & 235969 &  & 25436 &  \\
\rowgroup{Likelihood ratio (vs. null)} & ~ & 2427* & ~ & 3995* & ~ \\
\bottomrule
\end{tabular}
\caption{Regression results for predicting integrated verb use for loanwords. * indicates $p<0.01$, otherwise $p>0.01$; Bonferroni correction applied for significance testing for individual coefficients. \textbf{Bold} indicates variables for which effects are significant across both conditions and point in opposite directions.} 
\label{tab:integrated_verb_regression_combined}
\end{table*}

\subsubsection{Speaker-level Factors: Formality}
First, we find the following trends with respect to formality.

\paragraph{Post context matters} 
    Speakers tend to use the integrated form more often for native verbs when using hashtags ($\beta$=0.099) and less often for both loanwords and native verbs when using @-mentions ($\beta$=-0.087 loanwords, $\beta$=-0.050 native verbs).
    Prior work demonstrated a similar effect with nonstandard English words on Twitter~\citep{pavalanathan2015} and found that hashtags and @-mentions correlated with larger and smaller audience expectations.
    Since formal language is often expected with a larger audience~\citep{bell1984}, Spanish speakers may naturally choose the integrated verb forms to adapt to a larger potential audience.
    For post length, we find that longer posts tend to have integrated verbs more often for loanwords ($\beta$=0.051) and less often for native verbs ($\beta$=-0.046).
    This effect may be related to post content (e.g. including direct objects for loanword verbs) but it may also reflect inherent differences in the perceptions of loanwords and native verbs.

\subsubsection{Speaker-level factors: Background}

For loanword and native verb integration, we find the following trends with respect to speaker background.

\paragraph{Information sharing affects integration differently} 
    We find that the frequent URL-sharing speakers are more likely to use the integrated form for loanwords ($\beta$=0.024), and less likely to use the integrated form for native verbs ($\beta$=-0.015).
    If we assume that people who share more URLs are more interested in sharing new information~\citep{holton2014}, then these people may also be more likely to use formal verb forms for newer words (loanwords) and informal forms for older words (native verbs), due to the speakers' increased awareness of how new information should be treated.
    For RT sharing, we find that authors who frequently retweet others are more likely to use the integrated form of native verbs ($\beta$=0.025), which suggests that authors with more social ties (higher network embeddedness; cf. \citeauthor{milroy1985} \citeyear{milroy1985}) tend toward more standard language choices for frequently used words, i.e. native verbs.
    
\paragraph{Latin American authors prefer integration}
    For both word groups, Latin American authors use integrated verbs at a higher rate ($\beta$=0.228 for loanwords, $\beta$=0.133 for native verbs).
    Prior studies in World Englishes have found that dialects in post-colonial countries such as India sometimes adopt more linguistically conservative features~\citep{sharma2017}, which may be reflected in the higher rate of verb integration in Latin America (cf. conservative pronunciation in Latin American Spanish; \citeauthor{guy2014} \citeyear{guy2014}).
    In contrast, authors from Europe tend to use less verb integration ($\beta$=-0.367 for loanwords, $\beta$=-0.223 for native verbs), which suggests that using standard forms is less important for mainland Spain authors due to the dialect's relative prestige~\citep{hernandez2009}.
\paragraph{More integration for monolinguals}
    For loanwords, high-Spanish authors use integrated verbs at a higher rate than low-Spanish authors ($\beta$=0.589), and medium-Spanish authors use integrated verbs at a slightly higher rate ($\beta$=0.424).
    For native verbs, both high-Spanish and medium-Spanish authors use integrated verbs at a higher rate than low-Spanish authors ($\beta$=0.606 high-Spanish, $\beta$=0.687 medium-Spanish).
    Integrated verbs may be considered canonical and therefore more accessible for monolingual speakers, while light verbs could be more readily accessible to bilingual speakers who may default to simpler light verb constructions~\citep{gonzalez2011}.
    For example, the loanword phrase \emph{dar un like} may sound more natural to a bilingual speaker who is uncertain of the acceptability of \emph{likear}.

We note that for some of the variables such as post length and URL sharing, the effect direction for loanword integration is the opposite of the direction for native word integration.
The use of loanwords may bear a different social meaning for speakers as compared to native words (e.g. speakers consider loanwords to be newer in their vocabulary, \citeauthor{levendis2019} \citeyear{levendis2019}), which results in different effects on integration for the same social variable.
However, we leave more careful investigation of the differences between the word types for future work.
\section{Discussion}

We investigate the tendency for Spanish-speaking authors to use integrated verb forms for English loanwords, with a corpus of social media data augmented with speaker-level information.

The study provides a data set of loanwords and native words that linguists can use to investigate specific contexts of usage (e.g. in quotations, \citeauthor{iwasaki1994} \citeyear{iwasaki1994}).
The study also offers a pipeline for collecting various forms of loanwords using structured data (dictionaries) and data ``in the wild.''
More broadly, our work demonstrates the utility of social media as a window into speaker-level and contextual factors that underlie multilingual phenomena such as loanwords.

Our analyses show that integrated verb use for loanwords is clearly connected to underlying expectations of formality and standardness in language use, which also apply to native verbs.
The findings of this study provide additional context to prior work that showed some social correlates of loanword integration such as neighborhood composition~\citep{poplack1988}.
The decision to use integrated verb forms appears to rely not just on the speakers' background (e.g. linguistic knowledge) but even utterance-level context (e.g. audience), suggesting that the process is not ``inevitable''~\citep{poplack2012}.
Furthermore, the differences in domain-level and speaker-level effects across word groups (and within word groups, e.g. \autoref{tab:integrated_rate_differences_social_media_newspapers}) suggest different social perceptions, i.e. ``marked'' loanwords versus older, well-accepted native verbs.
Such implicit social evaluations can help predict the long-term entrenchment of loanwords in a speech community ~\cite{chesley2010,zenner2012}, and shed light on processes of cross-cultural contact and attitudes~\cite{levari2014experiment}.

This study has several limitations that merit further research.
First, the findings are narrowly focused on one form of integration, i.e. the alternation between different verb forms.
Future work should consider other forms of loanword integration on social media, including in orthography (Eng. \emph{football} $\rightarrow$ Sp. \emph{f\'utbol}) and syntax (\emph{\underline{el} key} vs. \emph{\underline{la} key} ``the key'')~\citep{montes2011,vendelin2006}.
It may be the case that some forms of loanword integration are more socially salient than others~\citep{myer1998} and therefore more strongly constrained by factors such as audience expectations.
In addition, this analysis found some location-level effects but did not zoom in to the level of the community, which is important since different speech communities may have different perceptions of the social value of loanwords~\citep{aaron2015,garley2014}.
As people of different linguistic backgrounds continue to interact on social media~\citep{kim2014}, it will be important to consider how different sub-communities on the platform adopt loanwords from one another, as such processes can lead to long-term language change.
Lastly, different languages may have different expectations about the social meaning of integrated loanword use, e.g. integrated verbs in Japanese may seem less formal than their light verb equivalent~\citep{tsujimura2011}.
More cross-linguistic work is needed to understand how well the social ramifications of loanword integration can be generalized~\citep{haspelmath2009} and whether they reflect culture-specific norms rather than inherent trends about language and socialization.
\section*{Acknowledgments}

This project was funded under NSF CAREER grant \#1452443 to JE and a Data Curation Award from Georgia Institute of Technology's Institute for Data Engineering and Science (IDEaS) to DY.
The authors thank Dr. Cecilia Montes-Alcal\'{a} and Dr. Lewis Chad Howe for their feedback on the validity of the loanword and native word pairs, as well as their feedback on early paper drafts.
The authors also thank members of the Computational Linguistics lab and the SALT Lab at Georgia Institute of Technology for their feedback.

\bibliography{main}

\begin{thebibliography}{48}
\expandafter\ifx\csname natexlab\endcsname\relax\def\natexlab#1{#1}\fi

\bibitem[{Aaron(2015)}]{aaron2015}
Jessi~Elana Aaron. 2015.
\newblock {Lone English-origin nouns in Spanish: The precedence of community
  norms}.
\newblock \emph{International Journal of Bilingualism}, 19(4):459--480.

\bibitem[{Bell(1984)}]{bell1984}
Allan Bell. 1984.
\newblock Language style as audience design.
\newblock \emph{Language in society}, 13(2):145--204.

\bibitem[{Biber and Conrad(2019)}]{biber2019}
Douglas Biber and Susan Conrad. 2019.
\newblock \emph{Register, genre, and style}.
\newblock Cambridge University Press.

\bibitem[{Boersma et~al.(2009)Boersma, Hamann et~al.}]{boersma2009}
Paul Boersma, Silke Hamann, et~al. 2009.
\newblock Loanword adaptation as first-language phonological perception.
\newblock \emph{Loanword phonology}, pages 11--58.

\bibitem[{Buckingham(2013)}]{buckingham2013}
Louisa Buckingham. 2013.
\newblock {Light verb constructions in Latin American newspapers: Creative
  variants and coinages}.
\newblock \emph{Spanish in Context}, 10(1):114--135.

\bibitem[{Cacoullos and Aaron(2003)}]{cacoullos2003}
Rena~Torres Cacoullos and Jessi~Elana Aaron. 2003.
\newblock {Bare English-origin nouns in Spanish: Rates, constraints, and
  discourse functions}.
\newblock \emph{Language Variation and Change}, 15(3):289--328.

\bibitem[{Chesley(2010)}]{chesley2010}
Paula Chesley. 2010.
\newblock {Lexical borrowings in French: Anglicisms as a separate phenomenon}.
\newblock \emph{Journal of French Language Studies}, 20(3):231--251.

\bibitem[{Chhaya et~al.(2018)Chhaya, Chawla, Goyal, Chanda, and
  Singh}]{chhaya2018}
Niyati Chhaya, Kushal Chawla, Tanya Goyal, Projjal Chanda, and Jaya Singh.
  2018.
\newblock Frustrated, polite, or formal: Quantifying feelings and tone in
  email.
\newblock In \emph{Proceedings of the Second Workshop on Computational Modeling
  of People’s Opinions, Personality, and Emotions in Social Media}, pages
  76--86.

\bibitem[{Coats(2018)}]{coats2018}
Steven Coats. 2018.
\newblock {Variation of New German Verbal Anglicisms in a Social Media Corpus}.
\newblock In \emph{Proceedings of the 6th conference on CMC and social media
  corpora for the humanities}.

\bibitem[{Do{\u{g}}ru{\"o}z and Nakov(2014)}]{dougruoz2014}
A.~Seza Do{\u{g}}ru{\"o}z and Preslav Nakov. 2014.
\newblock Predicting dialect variation in immigrant contexts using light verb
  constructions.
\newblock In \emph{EMNLP}, pages 1391--1395.

\bibitem[{Garley(2014)}]{garley2014}
Matt Garley. 2014.
\newblock Seen and not heard: The relationship of orthography, morphology, and
  phonology in loanword adaptation in the german hip hop community.
\newblock \emph{Discourse, Context \& Media}, 3:27--36.

\bibitem[{Garley and Hockenmaier(2012)}]{garley2012}
Matt Garley and Julia Hockenmaier. 2012.
\newblock {Beefmoves: dissemination, diversity, and dynamics of English
  borrowings in a German hip hop forum}.
\newblock In \emph{ACL}, pages 135--139.

\bibitem[{Geeraerts(2003)}]{geeraerts2003}
Dirk Geeraerts. 2003.
\newblock Cultural models of linguistic standardization.
\newblock In René Dirven, Roslyn Frank, and Martin Pütz, editors,
  \emph{Cognitive models in language and thought. Ideology, metaphors and
  meanings}, volume 2568.

\bibitem[{Gonz{\'a}lez-Vilbazo and L{\'o}pez(2011)}]{gonzalez2011}
Kay Gonz{\'a}lez-Vilbazo and Luis L{\'o}pez. 2011.
\newblock Some properties of light verbs in code-switching.
\newblock \emph{Lingua}, 121(5):832--850.

\bibitem[{Guy(2014)}]{guy2014}
Gregory Guy. 2014.
\newblock {Variation and change in Latin American Spanish and Portuguese}.
\newblock In \emph{Portuguese-Spanish interfaces: Diachrony, synchrony, and
  contact}, pages 443--464.

\bibitem[{Hall-Lew et~al.(2010)Hall-Lew, Coppock, and Starr}]{hall2010}
Lauren Hall-Lew, Elizabeth Coppock, and Rebecca~L Starr. 2010.
\newblock {Indexing political persuasion: Variation in the Iraq vowels}.
\newblock \emph{American Speech}, 85(1):91--102.

\bibitem[{Haspelmath(2009)}]{haspelmath2009}
Martin Haspelmath. 2009.
\newblock {Lexical borrowing: Concepts and issues}.
\newblock In \emph{Loanwords in the world's language: A Comparative Handbook},
  pages 944--967.

\bibitem[{Hern{\'a}ndez-Campoy and Villena-Ponsoda(2009)}]{hernandez2009}
Juan~Manuel Hern{\'a}ndez-Campoy and Juan~Andr{\'e}s Villena-Ponsoda. 2009.
\newblock {Standardness and nonstandardness in Spain: dialect attrition and
  revitalization of regional dialects of Spanish}.
\newblock \emph{International Journal of the Sociology of Language},
  2009(196-197):181--214.

\bibitem[{Holton et~al.(2014)Holton, Baek, Coddington, and
  Yaschur}]{holton2014}
Avery~E Holton, Kang Baek, Mark Coddington, and Carolyn Yaschur. 2014.
\newblock {Seeking and sharing: Motivations for linking on Twitter}.
\newblock \emph{Communication Research Reports}, 31(1):33--40.

\bibitem[{Iwasaki(1994)}]{iwasaki1994}
Yasufumi Iwasaki. 1994.
\newblock {Englishization of Japanese and acculturation of English to Japanese
  culture}.
\newblock \emph{World Englishes}, 13(2):261--272.

\bibitem[{Jenkins(2003)}]{jenkins2003}
Devin~L. Jenkins. 2003.
\newblock {Bilingual Verb Constructions in Southwestern Spanish}.
\newblock \emph{Bilingual Review}, pages 195--204.

\bibitem[{Kang(2011)}]{kang2011}
Yoonjung Kang. 2011.
\newblock Loanword phonology.
\newblock \emph{The Blackwell companion to phonology}, pages 1--25.

\bibitem[{Kariryaa et~al.(2018)Kariryaa, Johnson, Sch{\"o}ning, and
  Hecht}]{kariryaa2018}
Ankit Kariryaa, Isaac Johnson, Johannes Sch{\"o}ning, and Brent Hecht. 2018.
\newblock Defining and predicting the localness of volunteered geographic
  information using ground truth data.
\newblock In \emph{CHI}, pages 1--12.

\bibitem[{Kilgarriff(2010)}]{kilgarriff2010}
Adam Kilgarriff. 2010.
\newblock {Google the verb}.
\newblock \emph{Language Resources and Evaluation}, 44(3):281--290.

\bibitem[{Kim et~al.(2014)Kim, Weber, Wei, and Oh}]{kim2014}
Suin Kim, Ingmar Weber, Li~Wei, and Alice Oh. 2014.
\newblock {Sociolinguistic analysis of Twitter in multilingual societies}.
\newblock In \emph{{Proceedings of the 25th ACM conference on Hypertext and
  social media}}, pages 243--248.

\bibitem[{Lev-Ari and Peperkamp(2014)}]{levari2014experiment}
Shiri Lev-Ari and Sharon Peperkamp. 2014.
\newblock {An experimental study of the role of social factors in language
  change: The case of loanword adaptations}.
\newblock \emph{Laboratory Phonology}, 5(3):379--401.

\bibitem[{Levendis and Calude(2019)}]{levendis2019}
Katharine Levendis and Andreea Calude. 2019.
\newblock {Perception and Flagging of Loanwords--A diachronic case-study of
  M{\=a}ori loanwords in New Zealand English}.
\newblock \emph{Ampersand}, 6:100056.

\bibitem[{Lipski(1994)}]{lipski1994}
John Lipski. 1994.
\newblock \emph{Latin {American} {Spanish}}.
\newblock Longman, New York.

\bibitem[{Llopis and S{\'a}nchez-Lafuente(2009)}]{llopis2009}
Mar{\'\i}a {\'A}ngeles~Orts Llopis and {\'A}ngela~Almela S{\'a}nchez-Lafuente.
  2009.
\newblock {Translating the Spanish economic discourse of the crisis: Dealing
  with the inevitability of English loanwords}.
\newblock \emph{International Journal of English Studies}, 9(3):133--158.

\bibitem[{Lui and Baldwin(2012)}]{lui2012}
Marco Lui and Timothy Baldwin. 2012.
\newblock langid. py: An off-the-shelf language identification tool.
\newblock In \emph{ACL}, pages 25--30.

\bibitem[{Milroy and Milroy(1985)}]{milroy1985}
James Milroy and Lesley Milroy. 1985.
\newblock Linguistic change, social network and speaker innovation.
\newblock \emph{Journal of linguistics}, 21(2):339--384.

\bibitem[{Montes-Alcal{\'a} and Shin(2011)}]{montes2011}
Cecilia Montes-Alcal{\'a} and Naomi~Lapidus Shin. 2011.
\newblock Las keys versus el key: Feminine gender assignment in mixed-language
  texts.
\newblock \emph{Spanish in context}, 8(1):119--143.

\bibitem[{Myers-Scotton(1998)}]{myer1998}
Carol Myers-Scotton. 1998.
\newblock A theoretical introduction to the markedness model.
\newblock In \emph{{Codes and consequences: Choosing linguistic varieties}}.

\bibitem[{Pavalanathan and Eisenstein(2015)}]{pavalanathan2015}
Umashanthi Pavalanathan and Jacob Eisenstein. 2015.
\newblock Audience-modulated variation in online social media.
\newblock \emph{American Speech}, 90(2):187--213.

\bibitem[{Pavlick and Tetreault(2016)}]{pavlick2016}
Ellie Pavlick and Joel Tetreault. 2016.
\newblock An empirical analysis of formality in online communication.
\newblock \emph{Transactions of the Association for Computational Linguistics},
  4:61--74.

\bibitem[{Peperkamp(2004)}]{peperkamp2004}
Sharon Peperkamp. 2004.
\newblock A psycholinguistic theory of loanword adaptations.
\newblock In \emph{Annual Meeting of the Berkeley Linguistics Society},
  volume~30, pages 341--352.

\bibitem[{Poplack(1988)}]{poplack1988}
Shana Poplack. 1988.
\newblock Contrasting patterns of code-switching in two communities.
\newblock \emph{Codeswitching: Anthropological and sociolinguistic
  perspectives}, 48:215--244.

\bibitem[{Poplack and Dion(2012)}]{poplack2012}
Shana Poplack and Nathalie Dion. 2012.
\newblock Myths and facts about loanword development.
\newblock \emph{Language Variation and Change}, 24(3):279--315.

\bibitem[{Poplack et~al.(2020)Poplack, Robillard, Dion, and
  Paolillo}]{poplack2020}
Shana Poplack, Suzanne Robillard, Nathalie Dion, and John~C. Paolillo. 2020.
\newblock {Revisiting phonetic integration in bilingual borrowing}.
\newblock \emph{Language}, 96(1):126--159.

\bibitem[{Rodney and Jubilado(2012)}]{rodney2012}
C~Rodney and C~Jubilado. 2012.
\newblock {Morphological Study of Verb of Anglicisms in Spanish Computer
  Language}.
\newblock \emph{Polyglossia}, 23:43--47.

\bibitem[{Sanchez(2005)}]{sanchez2005}
Tara Sanchez. 2005.
\newblock The (socio-)linguistics of morphological borrowing: A quantitative
  look at qualitative constraints and universals.
\newblock \emph{University of Pennsylvania Working Papers in Linguistics},
  11(2):12.

\bibitem[{Sharma(2017)}]{sharma2017}
Devyani Sharma. 2017.
\newblock {English in India}.
\newblock In \emph{Varieties of English}, pages 311--329.

\bibitem[{Tsujimura and Davis(2011)}]{tsujimura2011}
Natsuko Tsujimura and Stuart Davis. 2011.
\newblock {A construction approach to innovative verbs in Japanese}.
\newblock \emph{Cognitive Linguistics}, 22(4):799--825.

\bibitem[{Vendelin and Peperkamp(2006)}]{vendelin2006}
Inga Vendelin and Sharon Peperkamp. 2006.
\newblock The influence of orthography on loanword adaptations.
\newblock \emph{Lingua}, 116(7):996--1007.

\bibitem[{Wang et~al.(2019)Wang, Hale, Adelani, Grabowicz, Hartman, Fl\"ock,
  and Jurgens}]{wang2019}
Zijian Wang, Scott Hale, David~Ifeoluwa Adelani, Przemyslaw Grabowicz, Timo
  Hartman, Fabian Fl\"ock, and David Jurgens. 2019.
\newblock Demographic inference and representative population estimates from
  multilingual social media data.
\newblock In \emph{The Web Conference}, pages 2056--2067.

\bibitem[{Wichmann and Wohlgemuth(2008)}]{wichmann2008}
S{\o}ren Wichmann and Jan Wohlgemuth. 2008.
\newblock Loan verbs in a typological perspective.
\newblock \emph{Empirical approaches to language typology}, 35:89.

\bibitem[{Wohlgemuth(2009)}]{wohlgemuth2009}
Jan Wohlgemuth. 2009.
\newblock \emph{A typology of verbal borrowings}, volume 211.
\newblock Walter de Gruyter.

\bibitem[{Zenner et~al.(2012)Zenner, Speelman, and Geeraerts}]{zenner2012}
Eline Zenner, Dirk Speelman, and Dirk Geeraerts. 2012.
\newblock {Cognitive Sociolinguistics meets loanword research: Measuring
  variation in the success of anglicisms in Dutch}.
\newblock \emph{Cognitive Linguistics}, 23(4):749--792.

\end{thebibliography}
\bibliographystyle{acl_natbib}

\appendix
\section{Appendix}
\label{sec:appendix}

\subsection{All integrated and light verb pairs}

To assist study replication, we list all pairs of integrated and light verbs for loanwords and native verbs used in this study.
We list them in alphabetical order (by integrated verb) in the format: \\ \emph{loanword/translation}: integrated verb ; light verb phrase(s)

\paragraph{Loanwords}
\begin{itemize}
    \small
    \setlength\itemsep{-5pt}
    \item \emph{access}: accesar ; hacer/tener acces
    \item \emph{aim}: aimear ; hacer/tener aim
    \item \emph{alert}: alertear ; hacer alert
    \item \emph{audit}: auditar ; hacer (un) audit
    \item \emph{ban}: banear ; hacer un ban
    \item \emph{bang}: bangear ; hacer bang
    \item \emph{bash}: bashear ; hacer/dar bash
    \item \emph{block}: blockear ; hacer/dar (un) block
    \item \emph{boycott}: boicotear ; hacer (un) boicot
    \item \emph{box}: boxear ; hacer (el) box/boxing
    \item \emph{bully}: bulear ; hacer/ser (el) bully
    \item \emph{bust}: bustear ; hacer (el) bust
    \item \emph{cast}: castear ; hacer cast/casting
    \item \emph{change}: changear ; hacer change
    \item \emph{chat}: chatear ; hacer chat
    \item \emph{check}: chequear ; hacer un cheque
    \item \emph{shoot}: chutar ; hacer/tomar el shot
    \item \emph{combat}: combatear ; hacer (el) combat
    \item \emph{connect}: conectar ; hacer (un) conexión
    \item \emph{crack}: crackear ; hacer crack
    \item \emph{customize}: customizar ; hacer custom/customized
    \item \emph{default}: defaultear ; hacer default
    \item \emph{delete}: deletear ; hacer/poner delete
    \item \emph{DM}: dmear ; mandar/enviar/poner un dm
    \item \emph{dope}: dopar ; hacer doping
    \item \emph{downvote}: downvotear ; poner/dar (un) downvote
    \item \emph{draft}: draftear ; hacer/tener draft
    \item \emph{drain}: drenar ; hacer (el) dren
    
    % \item \emph{encrypt}: encriptar ; hacer/ser encript
    \item \emph{smash}: esmachar ; hacer smash
    \item \emph{sniff}: esnifar ; hacer sniff
    \item \emph{standard}: estándar ; hacer (un) standard
    \item \emph{exit}: exitear ; hacer exit
    \item \emph{export}: exportear ; hacer export
    \item \emph{externalize}: externalizar ; hacer external
    \item \emph{fangirl}: fangirlear ; hacer/ser fangirl
    \item \emph{film}: filmar ; tomar (un) film
    \item \emph{flash}: flashear ; hacer (un) flash
    \item \emph{flex}: flexear ; hacer (un) flex
   
    \item \emph{flirt}: flirtear ; hacer flirt
    \item \emph{focus}: focalizar ; hacer focus
    \item \emph{format}: formatear ; hacer/dar (el) formato
    \item \emph{form}: formear ; hacer form
    \item \emph{freak}: friquear ; estar freaked
    \item \emph{freeze}: frizar ; hacer freeze
    \item \emph{fund}: fundear ; dar/hacer fund/funding
    \item \emph{gentrify}: gentrificar ; hacer/tener gentrificación
    \item \emph{ghost}: gostear ; hacer gost/ghost
    \item \emph{google}: googlear ; buscar en google
    \item \emph{hack}: hackear ; hacer hack
    \item \emph{hail}: hailear ; hacer hail
    \item \emph{hang}: hanguear ; hacer hang
    \item \emph{harm}: harmear ; hacer harm
    \item \emph{hypnosis}: hipnotizar ; hacer hipnosis
    \item \emph{host}: hostear ; hacer host
    \item \emph{hype}: hypear ; hacer hype
    \item \emph{intercept}: interceptear ; hacer/tirar interception
    \item \emph{hang}: janguear ; hacer hang (out)
    \item \emph{lag}: lagear ; hacer (un) lag
    \item \emph{like}: likear ; dar/poner (un) like
    \item \emph{limit}: limitear ; hacer (un) limit
    \item \emph{lynch}: linchar ; hacer lynch
    \item \emph{link}: linkear ; dar/poner (un) link
    \item \emph{love}: lovear ; hacer love
    \item \emph{look}: luquear ; dar/hacer (un) look
    \item \emph{make}: makear ; hacer make
    \item \emph{melt}: meltear ; hacer melt
    \item \emph{mope}: mopear ; hacer mope
    \item \emph{nag}: nagear ; hacer nag
    \item \emph{knock}: noquear ; dar/hacer (un) knockout
    \item \emph{pack}: packear ; hacer pack
    \item \emph{pan}: panear ; hacer/dar (un) panorama
    \item \emph{panic}: paniquear ; tener panic
    \item \emph{park}: parquear ; hacer parking
    \item \emph{perform}: performar ; hacer (un) performance
    \item \emph{pitch}: pichear ; hacer (un) pitch
    \item \emph{pin}: pinear ; hacer pin
    \item \emph{PM}: pmear ; enviar/mandar (un) pm
    \item \emph{punch}: ponchar ; hacer un punch
    \item \emph{post}: postear ; dar/poner (un) post
    \item \emph{posterize}: posterizar ; hacer poster
    \item \emph{print}: printear ; hacer print
    \item \emph{protest}: protestear ; hacer (un) protest
    \item \emph{push}: puchar ; hacer un push
    \item \emph{pump}: pumpear ; hacer pump(s)
    \item \emph{quote}: quotear ; hacer quote
    \item \emph{rank}: rankear ; hacer rank
    \item \emph{rant}: rantear ; hacer (un) rant
    \item \emph{rape}: rapear ; hacer (un) rape
    \item \emph{record}: recordear ; hacer (un) recording

    \item \emph{render}: renderizar ; hacer render(ed)
    \item \emph{rent}: rentear ; hacer rental/renting
    \item \emph{report}: reportear ; hacer (un) report
    \item \emph{reset}: resetear ; hacer reset
    \item \emph{respect}: respectear ; hacer respect
    \item \emph{ring}: ringear ; hacer ring
    \item \emph{rock}: rockear ; hacer rock
    \item \emph{roll}: rollear ; hacer roll
    \item \emph{sample}: samplear ; hacer (un) sample
    \item \emph{selfie}: selfiar ; tomar (un) selfie
    \item \emph{sext}: sextear ; dar/mandar un sext
    \item \emph{ship}: shippear ; hacer ship
    \item \emph{shitpost}: shitpostear ; hacer/poner un shitpost
    \item \emph{shock}: shockear ; hacer shock
    \item \emph{sign-in}: signear ; hacer sign-in
    \item \emph{stalk}: stalkear ; actuar como un stalker
    \item \emph{strike}: strikear ; hacer/dar un strike
    \item \emph{surf}: surfear ; hacer surf
    \item \emph{tackle}: taclear ; hacer tackle
    \item \emph{text}: textear ; mandar/enviar un text
    \item \emph{tick}: ticar ; hacer (un) tick
    \item \emph{torment}: tormentear ; hacer torment
    \item \emph{touch}: touchear ; hacer (un) touch
    \item \emph{transport}: transportear ; hacer transport
    \item \emph{travel}: travelear ; hacer travel
    \item \emph{troll}: trolear ; actuar como un trol
    \item \emph{tweet}: tweetear ; poner/enviar/hacer (un) tweet
    \item \emph{twerk}: twerkear ; hacer twerk
    \item \emph{upvote}: upvotear ; dar (un) upvote
    \item \emph{vape}: vapear ; hacer/tomar vape/vaping
    \item \emph{zap}: zapear ; hacer zap/zapping
\end{itemize}

\paragraph{Native verbs}
\begin{itemize}
    \small
    \setlength\itemsep{-5pt}
    \item \emph{admire}: admirar ; tener admiración
    \item \emph{befriend}: amistar ; tener amistad
    \item \emph{encourage}: animar ; subir el ánimo
    \item \emph{note}: anotar ; tomar nota
    \item \emph{land}: aterrizar ; hacer un aterrizaje
    \item \emph{joke}: bromear ; hacer bromas
    \item \emph{mock}: burlarse ; hacer burla
    \item \emph{punish}: castigar ; poner un castigo
    \item \emph{buy}: comprar ; hacer la compra
    \item \emph{copy}: copiar ; hacer una copia
    \item \emph{tickle}: cosquillar ; hacer cosquillas
    \item \emph{blame}: culpar ; echar la culpa
    \item \emph{damage}: dañar ; hacer daño
    \item \emph{decide}: decidir ; tomar decisiones
    \item \emph{apologize}: disculparse ; pedir disculpas
    \item \emph{shower}: ducharse ; darse una ducha
    \item \emph{question}: dudar ; poner en duda
    \item \emph{exemplify}: ejemplificar ; poner un ejemplo
    \item \emph{estimate}: estimar ; tener estima
    \item \emph{explain}: explicar ; dar una explicación
    \item \emph{finish}: finalizar ; poner fin
    \item \emph{photograph}: fotografiar ; tomar fotos
    \item \emph{escape}: fugarse ; darse a la fuga
    \item \emph{mention}: mencionar ; hacer mención
    \item \emph{look at}: mirar ; echar una mirada
    \item \emph{penalize}: multar ; poner una multa
    \item \emph{negotiate}: negociar ; hacer negocios
    \item \emph{originate}: originar ; dar origen
    \item \emph{participate}: participar ; tomar parte
    \item \emph{walk}: pasear ; dar un paseo
    \item \emph{step}: pisar ; poner el pie
    \item \emph{value}: preciar ; poner precio
    \item \emph{ask}: preguntar ; hacer (una) pregunta
    \item \emph{anticipate}: prever ; tener previsto
    \item \emph{test}: probar ; poner a prueba
    \item \emph{recommend}: recomendar ; hacer recomendación
    \item \emph{write}: redactar ; hacer una redacción
    \item \emph{cure}: remediar ; poner remedio
    \item \emph{breathe}: respirar ; dar un respiro
    \item \emph{jump}: saltar ; dar un salto
    \item \emph{nap}: sestear ; echar una siesta
    \item \emph{dream}: soñar ; tener un sueño
    \item \emph{end}: terminar ; poner término
    \item \emph{use}: usar ; hacer uso
    \item \emph{travel}: viajar ; hacer un viaje
    \item \emph{see}: vistar ; echar un vistazo
    \item \emph{fly}: volar ; tomar un vuelo
\end{itemize}

%

%
%

% \appendix

% \section{Appendices}
% \label{sec:appendix}

\end{document}